%% file: example_paper.tex
\documentclass{article}

\usepackage{microtype}
\usepackage{graphicx}
\usepackage{subfigure}
\usepackage{booktabs} % for professional tables

\usepackage{hyperref}

\usepackage[accepted]{icml2025}

\usepackage{amsmath}
\usepackage{amssymb}
\usepackage{mathtools}
\usepackage{amsthm}

\usepackage[capitalize,noabbrev]{cleveref}

\usepackage{graphicx}
\usepackage{subfigure}
\usepackage{amssymb,amsthm,dsfont,mathrsfs}
\usepackage{amsmath}
\usepackage{setspace}
\usepackage{multirow}
\usepackage{helvet}
\usepackage{courier}
\usepackage{bm}
\usepackage{mdwlist}
\usepackage{booktabs}
\usepackage{algorithm}
\usepackage{algorithmic}
\usepackage{url}

\theoremstyle{plain}

\theoremstyle{definition}

\theoremstyle{remark}

\usepackage[textsize=tiny]{todonotes}

\icmltitlerunning{Submission and Formatting Instructions for ICML 2025}

\begin{document}

\twocolumn[
\icmltitle{Self-Consistency of the Internal Reward Models Improves Self-Rewarding Language Models}

% \twocolumn[
% \icmltitle{Self-Consistent Internal Rewards Improve Self-Rewarding Language Models}




\begin{icmlauthorlist}
\icmlauthor{Xin Zhou}{yyy}
\icmlauthor{Yiwen Guo}{xxx}
\icmlauthor{Ruotian Ma}{yyy}
\icmlauthor{Tao Gui}{yyy}
\icmlauthor{Qi Zhang}{yyy}
\icmlauthor{Xuanjing Huang}{yyy}
% \icmlauthor{Firstname1 Lastname1}{equal,yyy}
% \icmlauthor{Firstname2 Lastname2}{equal,yyy,comp}
% \icmlauthor{Firstname3 Lastname3}{comp}
\end{icmlauthorlist}

\icmlaffiliation{yyy}{Fudan University}
\icmlaffiliation{xxx}{Independent Researcher}

\icmlcorrespondingauthor{Xin Zhou}{xzhou20@fudan.edu.cn}
\icmlcorrespondingauthor{Yiwen Guo}{guoyiwen89@gmail.com}
\icmlcorrespondingauthor{Qi Zhang}{qz@fudan.edu.cn}




\vskip 0.3in
]
% icmlEqualContribution


\printAffiliationsAndNotice{}  % leave blank if no need to mention equal contribution

\begin{abstract}
Aligning Large Language Models (LLMs) with human preferences is crucial for their deployment in real-world applications. 
Recent advancements in Self-Rewarding Language Models suggest that an LLM can use its internal reward models (such as LLM-as-a-Judge) \cite{yuanself} to generate preference data, improving alignment performance without costly human annotation. 
However, we find that different internal reward models within the same LLM often generate inconsistent preferences. 
This inconsistency raises concerns about the reliability of self-generated preference data, hinders overall alignment performance, and highlights the need for further research to ensure reliable and coherent alignment with human preferences.
To address this limitation, we propose Self-Consistent Internal Rewards (SCIR), a novel framework designed to enhance consistency among internal reward models during training. 
In each training step, we collect preference predictions from multiple pre-defined internal reward models and enforce consistency and confidence through an inconsistency penalty mechanism, thereby improving the reliability of these internal reward models. 
We selectively use data with consistent predictions for preference optimization, ensuring the quality of the preference data. 
By employing self-consistent internal rewards, our method significantly improves the alignment performance and reward modeling capability of LLMs, outperforming baseline methods by a notable margin.

% % ensuring the reliability of the internal reward models. 
% further avoids unreliable preference labels. 

\end{abstract}
\section{Introduction}
Large Language Models (LLMs) \cite{brown2020language, chowdhery2022palm, touvron2023llama} have demonstrated remarkable performance across various AI applications \cite{chatgpt, huang2023instruct2act, luo2023biomedgpt}. 
A crucial aspect of deploying Large Language Models (LLMs) in real-world scenarios is their alignment with human preferences \cite{bommasani2021opportunities}. This alignment is typically achieved through Reinforcement Learning from Human Feedback (RLHF) \cite{ouyang2022training, rafailov2024direct}, which trains LLMs to follow instructions and align with human preferences.
However, obtaining high-quality human-annotated preference data is costly and time-consuming, particularly when adapting to new domains or evolving requirements. 
Furthermore, the inherent limitations of human cognitive abilities can constrain the quality of preference data, presenting additional challenges for training super-intelligent AI  \cite{yuanself,burnsweak}.

Recent research has shown that the self-rewarding language model (SRLM) \cite{yuanself,wu2024metarewardinglanguagemodelsselfimproving,pang2024iterativereasoningpreferenceoptimization} is a promising approach to address these challenges. 
The core idea behind SRLM is to use LLMs themselves to generate preference data, reducing the need for human annotation or external reward models. 
In this paradigm, an LLM first acts as an instruction-following model to generate multiple responses, then serves as a reward model to evaluate the responses through LLM-as-a-Judge prompting \cite{zheng2023judgingllmasajudgemtbenchchatbot,gu2024surveyllmasajudge}, utilizing its generative reward modeling ability to provide preference data.
SRLM employs iterative DPO \cite{xu2024thingscringeothersiterative,pang2024iterative} to train the LLM on the self-generated preference data. Each training iteration not only enhances the LLM’s alignment performance but also improves its judging ability, providing better preferences data for the subsequent iteration. 

Despite its potential, we identify a limitation in the current SRLM paradigm: the preference labels predicted by the generative internal reward model (LLM-as-a-Judge) often conflicts with another internal reward model derived from the DPO training objective \cite{rafailov2024direct}. 
The inconsistency of internal reward models indicates that the preference data used in the SRLM process may not be reliable enough. 
Specifically, if the generative internal reward model is inaccurate, it indicates the model's LLM-as-a-Judge ability has not improved as expected, compromising the quality of preference data for the subsequent training iterations. 
Similarly, if the DPO-derived internal reward model is inaccurate, it implies that the DPO optimization direction and preference data in the previous training iterations are suboptimal. 
In both cases, such inconsistency could ultimately limit the overall alignment performance. 

In this paper, we propose imposing Self-Consistent Internal Rewards (SCIR) on the self-rewarding process. 
We argue that a reliable preference label should remain invariant across different reward models, and a well-aligned LLM should maintain self-consistency across its internal reward models. 
Specifically, SCIR uses two types of internal reward models: 
(1) the generative reward model, which instructs LLMs to generate preference judgments through carefully designed LLM-as-a-Judge prompts; and (2) the implicit reward model derived from DPO, which estimates rewards through the behavioral deviations from a reference model. 
In each training step, we use all internal reward models to predict preferences for each unlabeled preference pair, and apply an inconsistency penalty mechanism to make their predictions consistent and confident. 
We also design serval methods to mitigate the bias in the internal reward models. To further enhance the reliability of preference optimization, we only select the preference data with consistent predictions for DPO, using the model’s latest reward modeling ability to provide high-quality preference data.

To evaluate SCIR, we conduct experiments on Mistral-7B model series \cite{jiang2023mistral7b} , training the LLMs via  iterative DPO over three iterations. 
For each iteration, we randomly sample 4,000 prompts from the Stanford Alpaca Dataset \cite{alpaca} and generate preference data using the LLM alone, without human annotation or external reward models.
The experimental results demonstrate that our method effectively improves the model's alignment performance, achieving a 14\% improvement in length-controlled win rate on AlpacaEval 2.0, outperforming baselines. 
Additionally, we show that our approach improves the consistency of the LLMs' internal reward models and this consistency results in improved reward modeling ability. 

Our key contributions can be summarized as follows:
\begin{itemize}
    \item 
    We identify a limitation in SRLM: different internal reward models within the same LLM can generate inconsistent preferences. This inconsistency can hinder the quality of self-generated preference data and overall alignment performance.
    \item 
    We propose imposing Self-Consistent Internal Rewards (SCIR) on SRLM. SCIR improves the consistency of internal reward models via consistency loss, and uses dynamic preference optimization to ensure the quality of preference data, thereby improving performance.
    \item 
    Through comprehensive empirical evaluation, we demonstrate that SCIR can improve both alignment performance and reward modeling ability compared to the baselines, validating that self-consistency of internal reward models can improve SRLM.
    \end{itemize}

\section{Preliminaries}
\subsection{Direct Preference Optimization}
\label{sec:dpo}
Direct Preference Optimization (DPO) \cite{rafailov2024direct} is a widely used method for preference optimization. Unlike traditional approaches that rely on an explicit reward model, DPO reparameterizes the preference-based reward function using only the policy models:
\begin{equation}
    r(x, y) = \beta \log \frac{\pi_\theta(y \mid x)}{\pi_{\text{ref}}(y \mid x)} + \beta \log Z(x),
\end{equation}
where \(\pi_\theta\) is the policy model, \(\pi_{\text{ref}}\) is the reference policy, and \(Z(x)\) is the partition function. 
By integrating this reward formulation into the Bradley-Terry (BT) ranking model \cite{bradley1952rank}, DPO reformulates preference probabilities as:
\begin{equation}
    p(y_w \succ y_l \mid x) = \sigma \left( r(x, y_w) - r(x, y_l) \right),
\end{equation}
where $\sigma$ is the sigmoid function. 
This allows us to express preference probabilities directly with the policy model instead of relying on a separate reward model.
This implicit reward function also serves as the training objective of DPO: 
\begin{equation}
\small 
\begin{aligned}
\mathcal{L}_{\text{DPO}} (\pi_\theta; \pi_{\text{ref}}) = 
- \mathbb{E}_{(x, y_w, y_l) \sim \mathcal{D}} \Bigg[
    \log \sigma \Bigg(
        \beta \log \frac{\pi_\theta(y_w \mid x)}{\pi_{\text{ref}}(y_w \mid x)} \\
        - \beta \log \frac{\pi_\theta(y_l \mid x)}{\pi_{\text{ref}}(y_l \mid x)}
    \Bigg)
\Bigg],
\end{aligned}
\end{equation}
where \((x, y_w, y_l)\) consists of the prompt $x$, the winning response $y_w$, and the losing response $y_l$.

\subsection{Self-Rewarding Language Model} 
To address the problem of cost and quality in human-annotated preference data, SRLM \cite{yuanself} proposes using LLM to generate preference data by LLM-as-a-Judge prompting, achieving self-improvement without the need for human annotation or an external reward model.

In general, SRLM begins with a supervised fine-tuned (SFT) LLM $M_0$ and uses the iterative DPO \cite{xu2024thingscringeothersiterative,pang2024iterative} to train the model. $M_t$ refers to the model after $t$ training iterations.
Each iteration of SRLM consists of three steps.
\textbf{(1) Response Generation}: In the first step, the LLM should generate multiple response candidates for each given prompt. 
Formally, given the model $M_t$ and prompts$\{x_1,x_2,\dots,x_n\}$, $M_t$ generates $k$ responses $\{y_{i_1}, y_{i_2}, \dots, y_{i_k}\}$ for each prompt $x_i$. 
\textbf{(2) Preference Data Generation}: In the second step, SRLM uses the same LLM $M_t$ to generate preference data. $M_t$ is instructed to act as a judge and evaluate each response, assigning a score $r(y)$ to response $y$. Preference pairs are then constructed by selecting $(x, y_w, y_l)$ where $r(y_w) > r(y_l)$, forming the preference dataset $\mathcal{D}_t$ for this iteration.
\textbf{(3) Preference Optimization}: In the final step, the LLM $M_t$ is trained on preference data $\mathcal{D}_t$ using DPO, resulting in an improved model $M_{t+1}$. 

SRLM expects that each training iteration will simultaneously improve the model’s alignment performance and its LLM-as-a-Judge ability. In the subsequent iteration, the improved model $M_{t+1}$, with its enhanced LLM-as-a-Judge ability, will generate higher-quality preference data, leading to improved alignment performance.

\subsection{Internal Reward Model Inconsistency}
\label{sec:irm_inconsistency}
During SRLM training, we expect both the implicit DPO reward model to be well-optimized and the LLM's judgment capabilities to improve. 
Intuitively, a well-aligned LLM's DPO-derived reward model and LLM-as-a-Judge should provide a consistent and accurate preference label for the same preference pair. 
However, our empirical reveal substantial inconsistencies between these two internal reward models during the SRLM process. 
This inconsistency raises concerns about the alignment of LLM and the reliability of preference data in the SRLM. 

Following the SRLM setting proposed by \citet{yuanself}, we train Mistral-7B-v0.3 \cite{jiang2023mistral7b} through iterative DPO, using pointwise LLM-as-a-Judge to generate preference data. 
In each iteration $t$, we evaluate the internal reward models of the current model $M_t$ on two datasets: $\mathcal{D}_{t-1}$ (trained preference data from the previous iteration) and $\mathcal{D}_t$ (newly generated preference data). 
The inconsistency rate of internal reward models for each iteration is shown in Table \ref{tab:inconsistency}. 
We can observe that the LLM's two internal reward models generate inconsistent preference on approximately 50\% of samples across both datasets. 
As the number of iterations increases, the inconsistency rate on the trained data $\mathcal{D}_{t-1}$ slightly decreases, while the inconsistency rate on the new data $\mathcal{D}_{t}$ slightly increases. However, the overall inconsistency rate is still high across all iterations. 

We hypothesize that this inconsistency arises due to two reasons: 
\textbf{(1) Inaccurate preference data in the previous iteration.}
% (up to 95\%)
The DPO reward model is derived from the training objective, and its preference predictions on the trained data $\mathcal{D}_{t-1}$ are usually very close to the original preference labels, which are generated by previous model $M{t-1}$. 
Therefore, the observed inconsistencies on trained data  $\mathcal{D}_{t-1}$ primarily reflect disagreements between the LLM-as-a-Judge for current model $M_t$ and previous model $M{t-1}$.
As alignment performance improves, the enhanced LLM-as-a-Judge can provide better but inconsistent preference labels compared to previous iterations, 
suggesting that preference data used in the previous iterations are inaccurate and the optimization is not in the optimal direction. 
\textbf{(2) Limited improvement in LLM-as-a-Judge ability.} 
Despite improvements in alignment performance, the model's LLM-as-a-Judge ability may not improve as expected due to the absence of direct supervision during training. The suboptimal LLM-as-a-Judge can generate inconsistent and inaccurate preference data, which compromises the quality of preference data for DPO, limiting alignment performance in subsequent iterations. 
 \textbf{Therefore, the inconsistency of internal reward models presents an obstacle to achieving the optimal performance for SRLM.}

\input{tables/inconsistency_table}

\begin{figure*}[t]
\begin{center}
\includegraphics[width=1.0\linewidth]{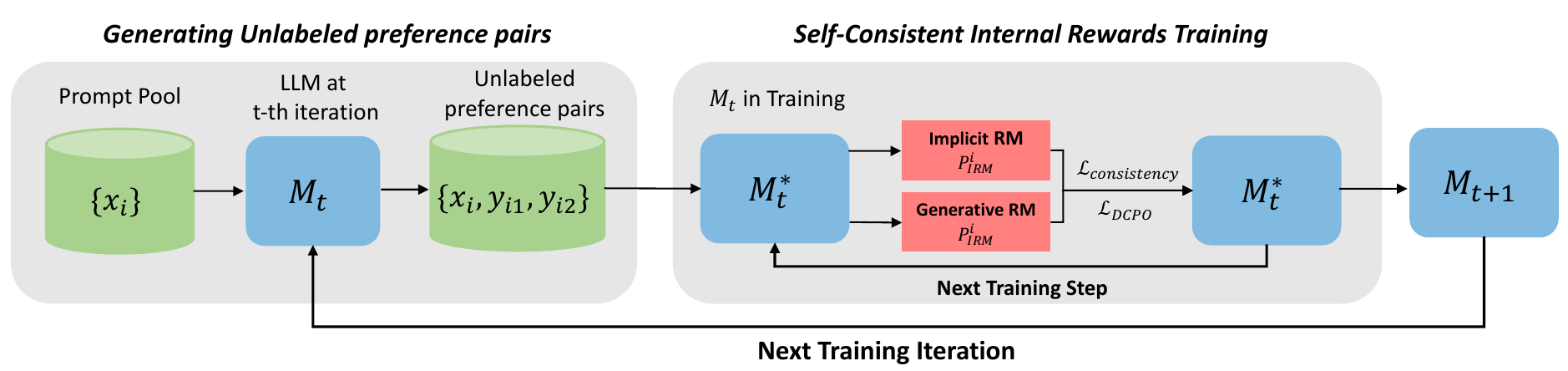}
\vskip -0.1in
\caption{An overview of our framework. 
For each iteration, the LLM $M_t$ generates responses for the prompts in the prompt pool, constructing unlabeled preference pairs. Then these pairs are used to optimize $M_t$ via Self-Consistent Intern Rewards (SCIR) Training. In each training step, the model's implicit DPO reward model and generative reward model predict the preference probabilities for each unlabeled preference pair. 
We use the consistency loss to encourage the preference probabilities of all internal reward models to be consistent. Meanwhile, preference pairs with consistent predictions across all internal reward models are selected for DPO optimization. 
The SCIR training results in model $M_{t+1}$, which is used for the next iteration.}
\label{fig:overview}
\end{center}
\vskip -0.1in
\end{figure*}

\section{Method}
In this section, we introduce Self-Consistent Internal Rewards (SCIR) to address the inconsistency of internal reward models and improve the overall alignment performance for SRLM. 
Figure \ref{fig:overview} provides an overview of our approach. SCIR consists of two key components: consistency training and dynamic consistency preference optimization (DCPO). The former improves internal reward model reliability by enhancing their consistency, while the latter ensures the quality of preference data by selectively choosing the data with consistent prediction from the latest internal reward models.

\subsection{Consistency Training}
Consistency Training aims to enhance the consistency of different internal reward models. 
Although our method is applicable to any internal reward model, this work focuses on two types of internal reward model:
a \textbf{generative reward model (GRM)} based on LLM-as-a-Judge prompting, and an \textbf{implicit reward model (IRM)} derived from DPO training objective, as shown in Section \ref{sec:dpo}. 

Given model $M_t$ and an unlabeled response pair $(x, y_1,y_2)$, we compute preference probabilities using all internal reward models of $M_t$. 
For the implicit reward model, the preference probability $P_{irm}(y_1 \succ y_2 \mid x)$ that IRM prefers $y_1$ over $y_2$ is: 
\begin{equation}
\label{eq:p_irm}
 \sigma \Bigg(
        \beta \log \frac{\pi_\theta(y_1 \mid x)}{\pi_{\text{ref}}(y_1 \mid x)} - \beta \log \frac{\pi_\theta(y_2 \mid x)}{\pi_{\text{ref}}(y_2 \mid x)}\Bigg),
\end{equation}
where $\sigma(\cdot)$ is the sigmoid function, $\pi_\theta$ is the $M_t$ and  $\pi_{\text{ref}}$ is the reference model. 

For generative reward model, we use pairwise LLM-as-a-Judge to avoid generation during training. 
The GRM preference probability $P_{\text{grm}}(y_1 \succ y_2 \mid x)$ is probability that LLM-as-a-Judge predicts $y_1$ is better:
\begin{equation}
P_{M_t}(y=\mathcal{V}(y_1)\mid x_{\text{Judge}},x,y_1,y_2)
\end{equation}
where $x_{\text{Judge}}$ is the judge prompt and $\mathcal{V}(y_1)$ is the pre-defined token indicating $y_1$ as the preferred response. 

For each training step, we use symmetrical Kullback-Leibler divergence \cite{zheng2021consistency}  to ensure consistency between these preference probabilities,  incorporating entropy regularization \cite{NIPS2004_96f2b50b,burnsdiscovering} and confidence-based masking \cite{xie2020unsupervised} to make the prediction confident.
Denoting $P$ as $P_{irm}$ and $Q$ as $P_{grm}$, the consistency training objective is:
\begin{equation}
\begin{aligned}
\label{eq:consistency}
\mathcal{L}_{\text{consistency}} = \mathbb{I}(P > \tau) \cdot \left[D_{\text{KL}}(Q \parallel \text{sg}(P)) + H(Q)\right] + \\
\mathbb{I}(Q > \tau) \cdot \left[D_{\text{KL}}(P \parallel \text{sg}(Q)) + H(P)\right],
\end{aligned}
\end{equation}
where $\mathbb{I}(\cdot)$ indicates the consistency loss applies only when the highest probability exceeds threshold $\tau$, and the \(\text{sg}(\cdot)\) operator stops gradients flowing during backpropagation, treating corresponding term as a fixed target. 
By ensuring consistency among all internal reward models, $\mathcal{L}_{\text{consistency}}$ identifies a reliable optimization direction for internal reward models, while simultaneously avoiding the trivial solution where $P_{dpo} = P_{\text{Judge}} = 0.5$ by making prediction confident.

\subsection{Dynamic Consistency Preference Optimization}
DCPO enhances preference data quality through two key mechanisms: utilizing the latest improved internal reward models to predict preference and selectively choosing the data with consistent prediction for DPO training. 
At each training step, all internal reward models predict the preference labels $\{r_1,r_2,\dots,r_n\}$ for 
each unlabeled preference pair, and only the data with consistent prediction are selected for DPO training.
The overall loss function of SCIR is:
\begin{equation}
\label{eq:dynamic_dpo}
\mathbb{I}(r_1=r_2=\dots=r_n) \cdot \mathcal{L}_{\text{DPO}} + \alpha \mathcal{L}_{\text{consistency}}
\end{equation}
where $\alpha$ is a hyperparameter that controls the strength of consistency loss and $r_i$ is the $i$-th internal reward model's preference label.
As the reward modeling ability improves during training, DCPO allows for a reliable and up-to-date reward signal, leading to enhanced alignment performance.  

\subsection{Additional Training Techniques}
\label{sec:additional_training}
We use additional techniques to address  some commonly encountered problems for internal reward models.
Existing research \cite{gu2025surveyllmasajudge} suggests that the LLM-as-a-Judge are sensitive to the choice of judge prompt templates and the position of responses within the prompt. 
To mitigate prompt bias and position bias, we employ two different judge prompt templates and alternate the positions of the responses within the judge prompts, thereby creating four different judge prompts. 
The average of the predictions from these four prompts is used as the preference prediction for the generative reward model. 
For the implicit reward model, we introduce an adaptive reference model technique to enhance its consistency, the details are shown in Appendix \ref{appendix:ada_ref}. 
The judge prompts are shown in the Appendix \ref{appendix:prompts}.

Besides that, the reward model tends to favor longer responses, and the internal reward model exhibits a similar issue \cite{singhal2023long,shen2023loose}. 
To alleviate this length bias, we take inspiration from \citet{park-etal-2024-disentangling} and introduce length regularization into the prediction logits of the internal reward model. 
Specifically, given the logits $\mathbf{o}$ that the internal reward model prefers a response $y$, the regularized logits are expressed as $\mathbf{o}' = \mathbf{o} - \alpha_l \cdot |y|$, where the hyperparameter $\alpha_l$ controls the strength of the regularization and $|y|$ is the length of response $y$.

\section{Experiments}
In this section, we first introduce the basic setup of our experiments in  Section \ref{sec:exp_setup}, and then show the experimental results of alignment in Section \ref{sec:exp_alignment_resuls}.
Section \ref{sec:exp_internal_reward_resuls} and Section \ref{sec:rm_ability} analyze the model's reward modeling ability. 
Finally, we conduct the ablation study in Section \ref{sec:ablation_study}.

\subsection{Experimental Setup} 
\label{sec:exp_setup}
\paragraph{Basic Setup}
Our experiments are primarily conducted on the Mistral-7b-v0.3 series \cite{jiang2023mistral7b}. 
To obtain a supervised fine-tuned LLM, we randomly sample 5,000 SFT examples from the Alpaca dataset \cite{alpaca} and combine them with the LIMA dataset \cite{zhou2023lima} to train Mistral-7b. 
We follow the setup in \citet{yuanself} and add 2,000 LLM-as-a-Judge examples to the SFT dataset, providing model the initial ability to act as a judge. 
In addition, we also fine-tune Mistral-7b-Instruct on LLM-as-a-Judge data to serve as a strong SFT model, validating the effectiveness of our method on both weak and strong initial LLMs. 
This fine-tuned model is referred to as $M_0$. 
Starting from $M_0$, we conduct three iterations of DPO training. The model trained after the $t$ iterations is denoted as \(M_{t}\). 
In each iteration, we randomly sample 4,000 prompts from Alpaca
dataset and use the current model to generate 4 candidate responses for each prompt. 
For our proposed SCIR, we randomly select two responses from these candidates to form an unlabeled preference pair, resulting in a total of 4,000 pairs as training data. 
These unlabeled pairs will be labeled during the SCIR training.
For other baselines, we use the corresponding reward model to generate preference data from candidate responses and train LLM on these preference data via DPO. 
We show other details in Appendix \ref{appendix:exp_setup}.

\paragraph{Baselines} 
We mainly compare our method with SRLM and its variants, their core difference is the reward model in the iterative DPO training phase.
\textbf{Self-Rewarding (LLM-as-a-Judge)} \cite{yuanself} is the standard SRLM method and uses a point-wise LLM-as-a-Judge paradigm to score each response. Preference data are constructed based on scores. 
\textbf{Self-Rewarding (Implicit Reward Model)} is a variant of SRLM that employs the implicit DPO reward model to generate preference data. 
Besides the SRLM, we also compare our method with \textbf{external reward model}, which uses an external reward model to predict preference labels. 
We chose the Skywork-Reward-8B \cite{liu2024skyworkrewardbagtricksreward} as the external baseline due to its outstanding reward modeling ability demonstrated across multiple benchmarks. 
\textbf{SCIR} is our proposed method, which uses consistency training and dynamic consistency preference optimization to enhance the SRLM.

\paragraph{Evaluation}
% We mainly evaluate the LLM's instruction following ability. 
We use widely adopted automatic evaluation datasets  AlpacaEval 2.0 \cite{dubois2024lengthcontrolledalpacaevalsimpleway} and MT-Bench \cite{zheng2023judging} to evaluate models' alignment performance.
\textbf{AlpacaEval 2.0} consists of 805 instructions chosen to be representative of user interactions in daily life. The evaluation metric is the win rate (WR) of the target model's responses compared to the responses of GPT4-Turbo, with GPT4-preview-1106 judges which response is better. 
Additionally, AlpacaEval 2.0 introduces the Length-Controlled Win Rate (LC) metric, which can reduce the bias of response length on the win rate.
\textbf{MT-Bench} uses 80 high-quality multi-turn questions to evaluate the LLMs' multi-turn conversational ability. It uses LLM-as-a-Judge to score the target model's responses to each turn conversation. We use GPT-4o as the judge and report the score for each turn conversation. 
We also utilize Massive Multitask Language Understanding (\textbf{MMLU}) \cite{hendrycksmeasuring} to evaluate the model's general knowledge and Grade School Math (\textbf{GSM8K}) \cite{cobbe2021training} to evaluate the model's reasoning ability.

\paragraph{Implementation Details}
We set hyperparameters based on preliminary experimental results and related works \cite{rafailov2024direct,yuanself,park-etal-2024-disentangling}. 
All of our experiments use the AdamW optimizer \cite{loshchilov2018decoupled} with cosine scheduling.
During the SFT phase,  we use a learning rate of 1e-6 and train the model for 5 epochs. 
For fine-tuning on the LLM-as-a-Judge data, we adjust the learning rate to 1e-7 and train for 1 epoch. 
For DPO training, we use a learning rate of 5e-7 and train for 2 epochs. 
The $\beta$ in the DPO loss is set to 0.1. 
For our proposed SCIR, we set $\tau=0.7$ in Equation \ref{eq:consistency} and $\alpha=1$ in Equation \ref{eq:dynamic_dpo}. 
The reference model in Equation \ref{eq:p_irm} is Mistral-7B. 
Length regularization is introduced in the second iteration of Mistral-7B-instruct and in the third iteration of Mistral-7B-v0.3, with $\alpha_l=0.02$ for GRM. 
We set the temperature to 0.7 and top-p to 0.9 for responses generation during both the iterative training and evaluation phases.

\input{tables/main_table}

\input{tables/instruct_table}

\subsection{Main Results}
\label{sec:exp_alignment_resuls}
% \paragraph{Overall Comparison}
% performs less effectively on weak LLMs compared to strong LLMs.
% % Similar phenomena are also observed in our experiments.  As shown in Table \ref{tab:instruct_results}, 
% performs less effectively compared to the strong SFT model.
We present the overall alignment performance in Table \ref{tab:main_results} and Table \ref{tab:instruct_results}.
From the results, we can find that:
\textbf{(1) Self-Rewarding (LLM-as-a-Judge)} achieves limited performance improvement. 
Previous work \cite{yuanself} has shown that SRLM with LLaMA-2-70B can achieve approximately 10\% win-rate (WR) improvement on Alpaca-Eval 2.0. Similar phenomena are also observed in Table \ref{tab:instruct_results}, SRLM with Mistral-7B-Instruct increases the WR of Alpaca-Eval2.0 from 16.14\% to 28.44\%. However, there is no such improvement observed in the more precise metrics Length-Controlled win-rate (LC). Instead, LC gradually decreases as the number of iterations increases, dropping from 26.74\% to 24.01\%. This suggests that SRLM may introduce a false improvement in alignment performance due to length bias. 
On the relatively weak SFT model Mistral-7B, the improvements from SRLM are limited, which suggests that SRLM's effectiveness may diminish on weaker initial model. We hypothesize that LLM-as-a-Judge alone may be insufficient to generate reliable preference data, especially for weak LLMs. 
We further investigate this hypothesis by evaluating SRLM's reward modeling ability in Section \ref{sec:rm_ability}. 
\textbf{(2) Self-Rewarding (Implicit Reward Model)} achieves better performance than Self-Rewarding (LLM-as-a-Judge). This suggests that the generative reward model may be less effective than the implicit reward model for Mistral-7B. Different internal reward models are suited to different scenarios. 
\textbf{(3) External Reward Model} outperforms SRLM baselines, aligning with expectations given its demonstrated strength in reward modeling. 
However, the external reward model is trained on external preference data, rather than on the models' outputs generated during the iterations. It may suffer from distribution shift \cite{dou2024metarmshifteddistributionsalignment,casper2023open}, limiting its overall performance. 
\textbf{(4) Our proposed SCIR} outperforms all baselines, including the external reward model. 
On Mistral-7B, SCIR increases LC win-rate of AlpacaEval 2.0 from 10.81 \% to 24.96\% and MT-Bench score from 5.39 to 6.18. 
SCIR with Mistral-7b-Instruct also achieves about 12\% improvement on LC win-rate. 
This success is attributed to consistency training and dynamic consistency preference optimization. 
By generating consistent and dynamically updated preference data, the model receives more reliable supervise signal for preference optimization, resulting in superior performance improvements. 
\textbf{(5) LLM's general abilities} do not largely change across all methods.
We observe minor fluctuations in performance on MMLU and GSM8K across iterations, but the overall changes are negligible. 
This is because we use a small learning rate during training and the training data is not closely aligned with tasks in MMLU and GSM8K, resulting in minimal impact on the model's general ability.

\subsection{Consistency of Internal Reward Models}
\label{sec:exp_internal_reward_resuls}

\begin{figure}[t]
% \vskip -0.2in
% \begin{center}
\centerline{\includegraphics[width=1\columnwidth]{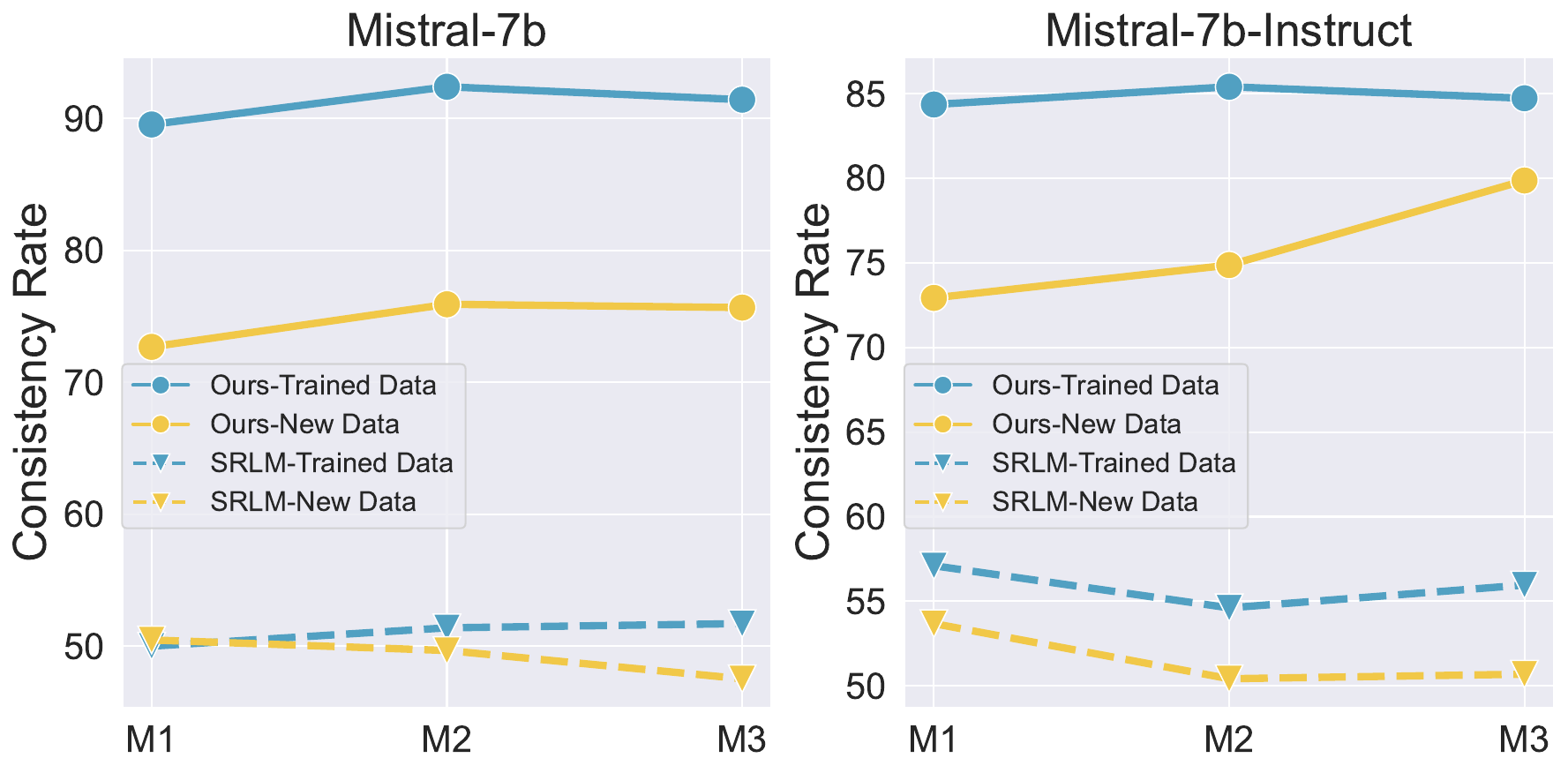}}
\caption{
Consistency rate of internal reward models. $M_t$ is the model after the $t$ iterations. New Data and Trained Data refer to the preference data from the $t$-th and the ($t$-1)-th iteration, respectively.}
\label{fig:inconsistency_exp}
% \end{center}
\vskip -0.15in
\end{figure}

% \paragraph{Consistency of Internal Reward Models} 
We follow the experimental setup in Section \ref{sec:irm_inconsistency}, using the model's internal reward models in $t$-th iteration to predict the labels for the preference data from both the current iteration (New Data $\mathcal{D}_{t}$) and previous iterations (Trained Data $\mathcal{D}_{t-1}$). 
The consistency rate is the proportion of cases where the generative and implicit reward models make consistent preference predictions on the same preference pairs.
Following Section \ref{sec:additional_training}, we take the predictions that are consistent across different judge prompts as the results of the generative reward model, and only use this subset of data to calculate the consistency rate. 
The experimental results of SRLM (LLM-as-a-Judge) and our method are shown in Figure \ref{fig:inconsistency_exp}. 
Similar to the previous results, SRLM's consistency rate on both types of data is not high. 
However, our method achieves a high consistency rate on both trained and new data, and the consistency rate continues to improve with increased iterations. 
This demonstrates that SCIR can effectively enhance the consistency of internal reward models.

% \paragraph{Reward Modeling Ability}
\subsection{Reward Modeling Ability}
\label{sec:rm_ability}
\paragraph{Experimental Setup}
We take the widely used reward model benchmark RewardBench \cite{lambert2024rewardbenchevaluatingrewardmodels} to evaluate the reward modeling ability of internal reward models. 
For each iteration, we use the internal reward models to predict the label of the preference pairs in RewardBench's Chat, Chat Hard, Safety, and Reasoning subsets. 
We directly use the average accuracy between the predicted results and the standard labels as the metric. 
It should be noted that some methods may only predict preference labels on subsets of the data. For example, SRLM uses a point-wise LLM-as-a-Judge, which may assign the same score to two responses, thus failing to predict preference labels correctly. 
Different methods may form preference data on different subsets, and these subsets are the actual data used in preference optimization, which can reflect the impact of preference data on the model training. 
As a result, we exclude the invalid pairs and only calculate the accuracy of data that could get valid preference labels. 
We show the accuracy of the GRM and IRM using different methods and additionally display the accuracy when the predictions of GRM and IRM are consistent (Consistency). 

\begin{figure}[t]
\centerline{\includegraphics[width=1\columnwidth]{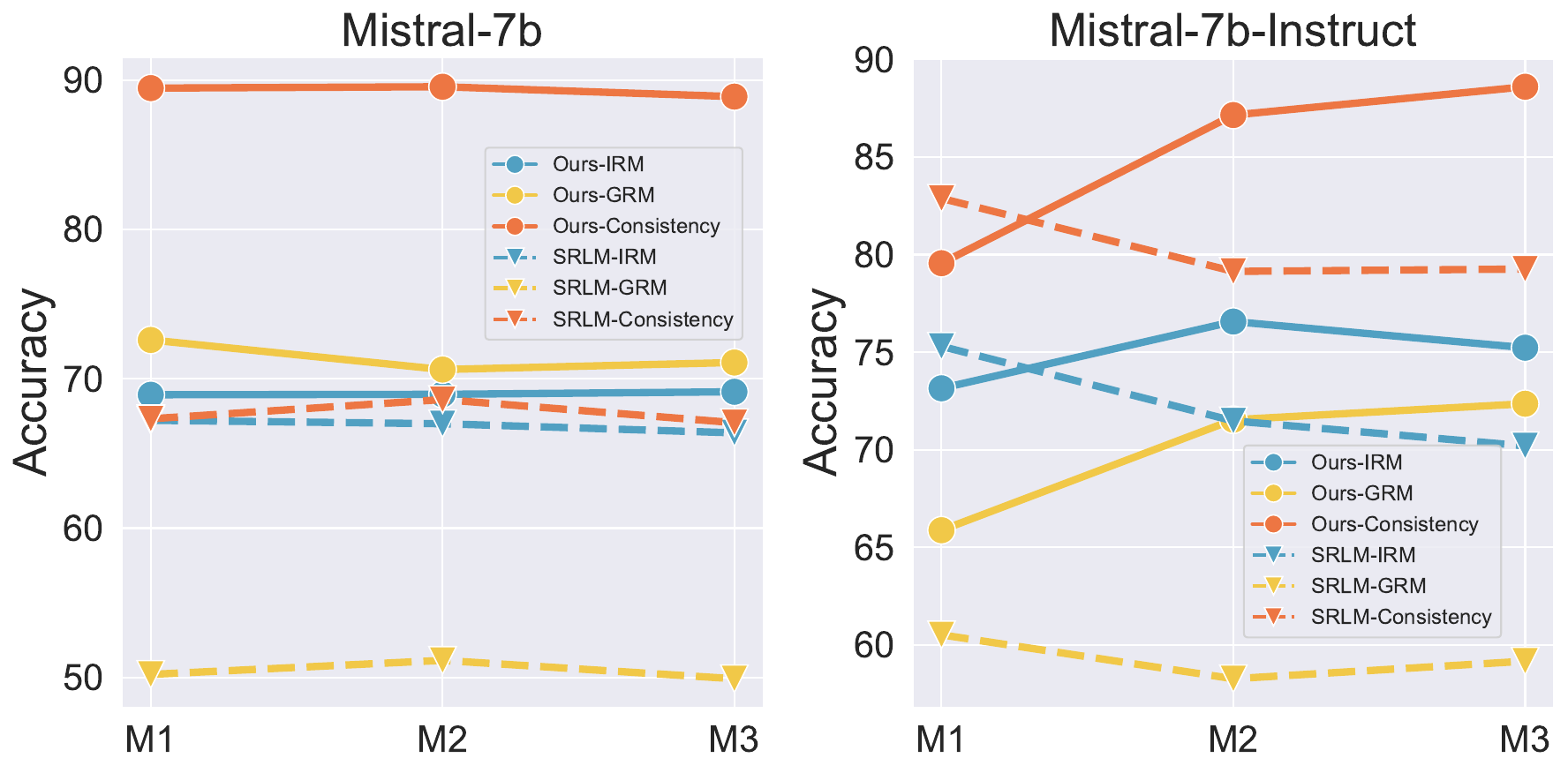}}
\caption{Results of the internal reward models on the subset of RewardBench. 
IRM is the implicit reward model and GRM is the generative reward model. 
Consistency means the IRM and GRM predict consistent preference labels. }
\label{fig:reward_modeling_exp}
\vskip -0.15in
\end{figure}

\paragraph{Results Analysis}
Results in Figure \ref{fig:reward_modeling_exp} reveals several key findings:
\textbf{First}, consistency between IRM and GRM leads to more reliable preferences. Both Ours-consistency and SRLM-consistency achieve higher accuracy than IRM or GRM alone, showing the importance of self-consistency. 
\textbf{Second}, stronger reward modeling ability often leads to better alignment performance.
For SRLM, GRMs of Mistral-7b and Mistral-7b-Instruct only achieve 50\% and 60\% accuracy. This insufficient reward modeling ability may explain why SRLM cannot effectively improve alignment performance, as discussed in Section \ref{sec:exp_alignment_resuls}. Moreover, the better alignment performance of Self-Rewarding (Implicit Reward Model) compared to Self-Rewarding (LLM-as-a-Judge) can also be attributed to IRM outperforming GRM.
\textbf{Third}, by improving consistency between IRM and GRM, Ours-consistency achieves superior accuracy on both types of LLMs, outperforming SRLM by a large margin. Additionally, on Mistral-7b-Instruct, the accuracy of Ours-consistency improves as the iteration increase, while the accuracy of SRLM-consistency without consistency training gradually decreases. We hypothesize that consistency training not only enhances consistency but also helps improve reward modeling ability. 
\textbf{Overall}, these results show that self-consistency of the internal reward models can enhance the reliability of intern reward models, and ultimately lead to improved alignment performance for SCIR. 

\subsection{Ablation Study}
\label{sec:ablation_study}

In this subsection, we perform ablation experiments to evaluate the contribution of SCIR's different loss function components.
Specifically, we start a new iteration from the M2 model of Mistral-7B, excluding various components: consistency training, dynamic preference optimization, multiple judge prompts, length regularization, and the adaptive reference model.
All other components and hyperparameters remain unchanged.
These ablated models are evaluated on AlpacaEval2.0, and the results are presented in Table \ref{tab:ablation}. 
As expected, Full SCIR achieves the best alignment performance. 
When multiple judge prompts, length regularization, and adaptive reference model are removed, the reliability of the preference data decreases, leading to a drop in alignment performance. 
Consistency training optimizes the IRM, which is inherently optimizing the DPO objective, thus improving performance even without dynamic preference optimization.
However, the model's preference optimization heavily relies on consistency training and dynamic consistency preference optimization. The absence of these two components leads to the most notable performance degradation. 
Overall, each component of SCIR contributes positively to alignment performance.

\input{tables/ablation_table}

\section{Related Work}
\textbf{Alignment of Large Language Model} is one key factor behind the LLMs' success \cite{bommasani2021opportunities,wang2023aligninglargelanguagemodels}. 
By aligning model behavior with human preferences, an LLM can follow human instructions and generate helpful and harmless responses \cite{bai2022traininghelpfulharmlessassistant}.  
Alignment performance relies on preference data and preference learning algorithms. 
A classic preference learning algorithm is Reinforcement Learning from Human Feedback (RLHF) \cite{bai2022training,bai2022constitutional,ouyang2022training}, which typically uses preference data to train an external reward model to score LLM's response, and then uses Proximal Policy Optimization (PPO) \cite{schulman2017proximalpolicyoptimizationalgorithms} to optimize the LLM, making LLMs' responses maximize the reward model’s score. 
Another widely used preference learning algorithm is Direct Preference Optimization (DPO) \cite{rafailov2024direct}, which modifies the optimization objective of RLHF, allowing the model to be trained directly on preference data without the need for a reward model during the training process. 
Both RLHF and DPO relies on humans or an additional reward model to annotate preference data. 
High-quality preference data can enhance the effectiveness of preference optimization, but collecting such data is often time-consuming and labor-intensive. 
Therefore, improving the quality of preference data and reducing the cost of collecting data are two promising directions for LLM alignment \cite{kaufmann2024surveyreinforcementlearninghuman,casper2023openproblemsfundamentallimitations}.

\textbf{Self-Rewarding Language Model} (SRLM) \cite{yuanself} has proposed that an LLM can generate preference data by itself. Training model on self-generated data can further enhance its alignment performance. 
SRLM provides a solution to avoid time-consuming and labor-intensive human preference annotation and reward model training \cite{bai2022training}, enabling rapid adaptation to new domains in a self-improvement manner \cite{huang2022largelanguagemodelsselfimprove}. 
Additionally, SRLM can help achieve superhuman agents. When an LLM exceeds human in a specific domain, it can autonomously provide superhuman feedback to ensure an adequate training signal \cite{burns2023weak}. 
To improve SRLM, \citet{anonymous2024metarewarding} 
suggests using the same LLM as a meta-judge to evaluate its generated LLM-as-a-Judge, thereby improving the ability of generative reward models. \citet{wang2024creamconsistencyregularizedselfrewarding} introduces the use of regularization to enhance the consistency of DPO rewards across different iterations, thus providing more robust preference data. 
While all these methods aim to improve the quality of preference data in SRLMs, this paper explores different strategies: enhancing consistency across different internal reward models within the same iteration to improve the reliability of both the internal reward models and the preference data.

\section{Conclusion}
In this paper, we explore the consistency of internal reward models for the self-rewarding language models. 
We reveal that during the process of SRLM, there is a inconsistency between the generative reward model (LLM-as-a-Judge) and the implicit reward model (DPO training objective). 
This inconsistency suggests that the internal reward model and the preference data during SRLM training may be unreliable. 
To mitigate this issue, we propose Self-Consistency Internal Rewards Training to promote the consistency of the internal reward model. 
During training, we use different internal reward models to predict the preference probabilities of unlabeled preference pairs, and apply a consistency loss to make these predictions consistent and confident. 
To enhance the reliability of preference optimization, we only select preference pairs that get consistent predictions across all internal reward models for DPO training. 
Experimental results demonstrate that our method significantly improves the model's alignment performance, surpassing baseline methods. Moreover, our analysis shows that enhancing the consistency of the internal reward model and selecting consistent preference data effectively boosts the model's reward modeling performance.

\section*{Impact Statement}
This paper presents work whose goal is to advance the field of Machine Learning. There are many potential societal consequences of our work, none which we feel must be specifically highlighted here.

\bibliography{example_paper}
\bibliographystyle{icml2025}

%%%%%%%%%%%%%%%%%%%%%%%%%%%%%%%%%%%%%%%%%%%%%%%%%%%%%%%%%%%%%%%%%%%%%%%%%%%%%%%
%%%%%%%%%%%%%%%%%%%%%%%%%%%%%%%%%%%%%%%%%%%%%%%%%%%%%%%%%%%%%%%%%%%%%%%%%%%%%%%
% APPENDIX
%%%%%%%%%%%%%%%%%%%%%%%%%%%%%%%%%%%%%%%%%%%%%%%%%%%%%%%%%%%%%%%%%%%%%%%%%%%%%%%
%%%%%%%%%%%%%%%%%%%%%%%%%%%%%%%%%%%%%%%%%%%%%%%%%%%%%%%%%%%%%%%%%%%%%%%%%%%%%%%
\newpage
\appendix
\onecolumn
\section{Adaptive Reference Model for Implicit Reward Model}
\label{appendix:ada_ref}
In Section \ref{sec:additional_training}, we introduce adaptive reference model to enhance the consistency of implicit reward model (IRM). In this section, we provide the details of the method.

We start from describing the implicit reward model, which is also the DPO training objective:
\begin{equation}
     \sigma \Bigg(
        \beta \log \frac{\pi_\theta(y_1 \mid x)}{\pi_{\text{ref}}(y_1 \mid x)} - \beta \log \frac{\pi_\theta(y_2 \mid x)}{\pi_{\text{ref}}(y_2 \mid x)}\Bigg),
\end{equation}
where the $\pi_\theta$ is the $M_t$ during the SRLM iterations, $\pi_{\text{ref}}$ is the reference model. 
In the standard DPO loss, $\pi_{\text{ref}}$ and $\pi_\theta$ are initially identical, but $\pi_{\text{ref}}$ is frozen and does not update parameter. We refer to the reference model used in the standard DPO method as the local reference model \( \pi_{\text{ref}}^l \).
When computing the IRM, we use the base model, such as Mistral-7B, as the reference model. This ensures that the impact of overall SRLM training iteration can be reflected in the IRM. We refer to the reference model used in IRM as the global reference model \( \pi_{\text{ref}}^g \).
Obviously, IRM with different reference model (\( \pi_{\text{ref}}^l \) or \( \pi_{\text{ref}}^g \)) may predict inconsistent preference labels. 

% We refer to the reference model used in the standard DPO method as the *local reference model* \( \pi_{\text{ref}}^l \), and the reference model used in IRM as the *global reference model* \( \pi_{\text{ref}}^g \). 

To address this issue, we propose the method called adaptive reference model. The goal is to ensure that the preference predictions made by \( \pi_{\text{ref}}^l \) and \( \pi_{\text{ref}}^g \) are consistent. 
Suppose that the IRM prefers \( y_1 \), which implies that the preference probability \( P_{\text{irm}}(y_1 \succ y_2 \mid x) > 0.5 \). This can be rewritten as:
\[
\beta \log \frac{\pi_\theta(y_1 \mid x)}{\pi_{\text{ref}}(y_1 \mid x)} - \beta \log \frac{\pi_\theta(y_2 \mid x)}{\pi_{\text{ref}}(y_2 \mid x)} > 0,
\]

which simplifies to:

\[
\beta \log \frac{\pi_\theta(y_1 \mid x)}{\pi_\theta(y_2 \mid x)} - \beta \log \frac{\pi_{\text{ref}}(y_1 \mid x)}{\pi_{\text{ref}}(y_2 \mid x)} > 0.
\]

$\beta$ is a hyperparameter greater than 0. Dividing through by $\beta$, we get:

\[
\frac{\pi_\theta(y_1 \mid x)}{\pi_\theta(y_2 \mid x)} > \frac{\pi_{\text{ref}}(y_1 \mid x)}{\pi_{\text{ref}}(y_2 \mid x)}.
\]
During training, only \( \pi_\theta \) can be updated, while \( \pi_{\text{ref}} \) remains fixed once set. 
As the result, to ensure consistent predictions with IRM using different reference models, we propose that \( \frac{\pi_\theta(y_1 \mid x)}{\pi_\theta(y_2 \mid x)} \) should always be greater than the maximum value of \( \frac{\pi_{\text{ref}}(y_1 \mid x)}{\pi_{\text{ref}}(y_2 \mid x)} \) across all reference models. Formally, we require:
\[
\frac{\pi_\theta(y_1 \mid x)}{\pi_\theta(y_2 \mid x)} > \max \left\{ \frac{\pi_{\text{ref}}^l(y_1 \mid x)}{\pi_{\text{ref}}^l(y_2 \mid x)}, \frac{\pi_{\text{ref}}^g(y_1 \mid x)}{\pi_{\text{ref}}^g(y_2 \mid x)} \right\}.
\]

Thus, we select the reference model—either the local reference model \( \pi_{\text{ref}}^l \) or the global reference model \( \pi_{\text{ref}}^g \)—based on which one maximizes the value of \( \frac{\pi_{\text{ref}}(y_1 \mid x)}{\pi_{\text{ref}}(y_2 \mid x)} \). This reference model is then used in the DPO loss function in Equation \ref{eq:dynamic_dpo}. Note that the global reference model \( \pi_{\text{ref}}^g \) is used for IRM in Equation \ref{eq:p_irm}, local reference model  \( \pi_{\text{ref}}^l \) is used for DPO loss in Equation \ref{eq:dynamic_dpo}. Adaptive reference model does not require loading additional model, which makes training efficient.
The inconsistency of IRM also improves the reliability of preference data and contribute to the overall alignment performance. The results of ablation study are shown in Table \ref{tab:ablation} in Section \ref{sec:ablation_study}.

Recently, CREAM \cite{wang2024creamconsistencyregularizedselfrewarding} also propose use consistency regularization to reduce the noise caused by variations in the reference models during training iterations. But CREAM primarily selects consistent preference data by utilizing the IRM with a local reference model from the $t$-th and $(t-1)$-th iterations. Our proposed adaptive reference model, on the other hand, uses $M_t$ and the base model as reference models to enhance the consistency of the IRM. This is because the goal of SCIR is to improve the consistency of all internal reward models. 
There are much differences in both the methods and motivations between the two approaches.

\section{Prompts for LLM-as-a-Judge}
\label{appendix:prompts}
In this section, we show the prompts for LLM-as-a-Judge used in our proposed SCIR and baselines. 
For our proposed SCIR, as mentioned in Section \ref{sec:additional_training}, we use two types of judge prompts for SCIR. These two prompts are shown in Table \ref{tab:judge_prompts_scir}. Judge Prompt 1 is the prompt used in AlpacaEval \cite{dubois2024lengthcontrolledalpacaevalsimpleway} and Judge Prompt 2 is the prompt used in Chatbot Arena \cite{zheng2023judging}. These judge prompt are widely used in LLM-as-a-Judge and can effectively select the better responses. 
For Self-Rewarding Language Model, we follow the \citet{yuanself} and directly use its judge prompt, which is shown in Table \ref{tab:judge_prompts_srlm}. 
\input{tables/SCIR_prompt}
\input{tables/SRLM_prompt}

\section{Details of Experimental Setup}
\label{appendix:exp_setup}
\subsection{Supervised Fine-tuneing}
In our main experiments, SFT dataset consists of 5,000 examples from Alpaca dataset \cite{alpaca}, 1,030 examples from LIMA \cite{zhou2023lima} dataset and 2,000 LLM-as-a-Judge data. Different baselines require different types of LLM-as-a-Judge data. We use preference version \footnote{\url{https://huggingface.co/datasets/monology/oasst2\_dpo}} of Oasst2 dataset \cite{kopf2024openassistant} to generate LLM-as-a-Judge data. 
\textbf{Our proposed SCIR} uses the pairwise LLM-as-a-Judge paradigm, where the goal is to select the better response between two responses. 
We randomly sample 500 examples from Oasst2 and applied the multiple judge templates. Each preference pair generates 4 LLM-as-a-Judge examples, resulting in a total of 2000 pairwise LLM-as-a-Judge data.
\textbf{Self-Rewarding (LLM-as-a-Judge)} use the pointwise LLM-as-a-Judge paradigm, where the goal is to score each response. 
We use LLama3-70B-Instruct \cite{grattafiori2024llama3herdmodels} as the LLM-as-a-Judge to generate the score of chosen and rejected responses. If the score of the chosen response is higher than that of the rejected response, we keep the two LLM-as-a-Judge examples for these two responses. 
We also generate 2000 pointwise LLM-as-a-Judge data to match the data used in SCIR. 
\textbf{Other baselines} do not use the  LLM-as-a-Judge data for training, because they do not use the LLM-as-a-Judge to generate preference data.

\subsection{Iterative Training}
We implement all the baselines ourselves due to the lack of open-source code and dataset.
In each iteration, the model generates 4 responses for each of the 5000 prompts from the Alpaca dataset.
For the Self-Rewarding(LLM-as-a-Judge), we use pointwise LLM-as-a-Judge to score each response and form preference pairs from the highest and lowest-scoring responses. Since different responses may have the same score, the number of preference pairs used in each iteration may be different. For Mistral-7b, 
the three iterations use 3696, 3588, and 3646 preference data, respectively. 
For Mistral-7b-Instruct, it generate 3075, 2957, and 3168 preference data, respectively.
For the Self-Rewarding(Implicit Reward Model) and external reward model, we use IRM or Skywork-reward-7B to assign rewards to each response and form preference pairs from the highest and lowest-scoring responses. Each iteration uses 4000 preference pairs.
For SCIR, we randomly select two responses to form an unlabeled preference pair, generating a total of 4000 unlabeled preference pairs for training.

\end{document}

%% file: tables/inconsistency_table.tex
% \begin{table}[t]
% \caption{Classification accuracies for naive Bayes and flexible
% Bayes on various data sets.}
% \label{sample-table}
% \vskip 0.15in
% \begin{center}
% \begin{small}
% \begin{sc}
% \begin{tabular}{lcccr}
% \toprule
% Data set & Naive & Flexible & Better? \\
% \midrule
% Breast    & 95.9$\pm$ 0.2& 96.7$\pm$ 0.2& $\surd$ \\
% Cleveland & 83.3$\pm$ 0.6& 80.0$\pm$ 0.6& $\times$\\
% Glass2    & 61.9$\pm$ 1.4& 83.8$\pm$ 0.7& $\surd$ \\
% Credit    & 74.8$\pm$ 0.5& 78.3$\pm$ 0.6&         \\
% Horse     & 73.3$\pm$ 0.9& 69.7$\pm$ 1.0& $\times$\\
% Meta      & 67.1$\pm$ 0.6& 76.5$\pm$ 0.5& $\surd$ \\
% Pima      & 75.1$\pm$ 0.6& 73.9$\pm$ 0.5&         \\
% Vehicle   & 44.9$\pm$ 0.6& 61.5$\pm$ 0.4& $\surd$ \\
% \bottomrule
% \end{tabular}
% \end{sc}
% \end{small}
% \end{center}
% \vskip -0.1in
% \end{table}

\begin{table}[t]
\caption{Inconsistency rate of internal reward models during SRLM iterations. $D_{t-1}$ is the preference data from previous iteration and $D_{t}$ is the new preference data for the current iteration.}
\label{tab:inconsistency}
% \vskip 0.15in
 \begin{center}
    \begin{tabular}{lcc}
        \toprule 
        Model & Trained Data ($\mathcal{D}_{t-1}$) & New Data ($\mathcal{D}_{t}$) \\ 
        \midrule
        Iteration 1 & 50.00\% & 49.53\% \\
        Iteration 2 & 48.61\% & 50.33\% \\
        Iteration 3 & 48.29\% &  52.47\%\\
        \bottomrule
    \end{tabular}
\end{center}
\vskip -0.1in
\end{table}

%% file: tables/main_table.tex
\begin{table*}[t!]
\centering
\small
\caption{Experimental results on Mistral-7B.  $M_0$ is the supervised fine-tuned LLM. $M_1$, $M_2$ and $M_3$ are the model after $i$ iterations of training. The best performance among all models is marked in bold.}
\label{tab:main_results}
\vskip 0.15in
    % \resizebox{0.8\linewidth}{!}{
    \begin{tabular}{lccccccccccccccccc}
    \toprule 
    % Arena-Hard
    \multirow{2}{*}{\textbf{Model}}  &  \multicolumn{3}{c}{\textbf{Alpaca-Eval 2.0}} & \multicolumn{3}{c}{\textbf{MT-Bench}} & \textbf{MMLU} & \textbf{GSM8K} \\
    % \cline{2-4} 
    & \textbf{LC} & \textbf{WR} & \textbf{Length} & \textbf{Turn 1} & \textbf{Turn 2} & \textbf{Score} & \textbf{Acc.} & \textbf{EM} \\
    \midrule
    \multicolumn{9}{c}{\textbf{\textit{Self-Rewarding (LLM-as-Judge)}}} \\
    % Self-Rewarding (LLM-as-Judge) \\
    $M_0$ & 10.65 & 6.35 & 965 & 6.68 & 4.68 & 5.68 & 59.61 & 35.41 \\
    $M_1$ & 12.58 & 7.33 & 1027 & 6.63 & 4.42 & 5.53  & 59.51 & 35.03\\
    $M_2$ & 11.37 & 7.58 & 1125 & 6.67 & 4.55 & 5.61 & 59.41 & 35.78 \\
    $M_3$ & 10.74 & 7.22 & 1192 & 6.17 & 4.92 & 5.55 & 59.30 & 35.63 \\
    \midrule
    \multicolumn{9}{c}{\textbf{\textit{Self-Rewarding (Implicit Reward Model)}}} \\
    % Self-Rewarding (Implicit Reward Model) \\
    $M_0$ & 10.66 & 6.35 & 952  & 6.62 & 4.55 & 5.58 & 59.65 & 37.45 \\
    $M_1$ & 10.43 & 7.08 & 1097  & 6.45 & 4.73 & 5.59 & 59.72 & 37.76 \\
    $M_2$ & 12.52 & 8.82 & 1218  &  6.43 & 4.98 & 5.70 & 59.76 & 37.45 \\
    $M_3$ & 14.15 & 11.43 & 1390  & 6.82 & 5.01 & 5.91 & 59.61 & \textbf{37.68}\\
    \midrule
    \multicolumn{9}{c}{\textbf{\textit{External Reward Model (Skywork-reward-7B)}}} \\
    % External Reward Model (Skywork-reward-7B) \\
    $M_0$ & 10.18 & 5.89 & 945 &  6.36 & 4.47 & 5.41 & 59.34 & 37.23 \\
    $M_1$ & 12.09 & 6.45 & 1147 & 6.73 & 5.13 & 5.93 & 59.33 & 36.32\\
    $M_2$ & 14.09 & 9.78 & 1271   & 6.71 & 4.93 & 5.82 & 59.35 & 36.24\\
    $M_3$ & 15.91 & 10.68 & 1291 & 6.75 & 4.85 & 5.80 & 59.39 & 35.10\\
    \midrule
    \multicolumn{9}{c}{\textbf{\textit{Self-Consistent Internal Rewards (Ours)}}} \\
    % Self-Consistent Internal Rewards (Ours) \\
    $M_0$ & 10.81 & 5.96 & 953 & 6.43 & 4.35 & 5.39 & \textbf{59.96} & 36.62\\
    $M_1$ & 14.51 & 11.06 & 1332 & \textbf{6.82} & 5.28 & 6.05 & 59.92 &  36.32\\
    $M_2$ & 18.49 & 17.02 & 1753 & 6.80 & 5.32 & 6.06 & 59.78 & 36.32\\
    $M_3$ & \textbf{24.92} & \textbf{23.85} & 1941 & 6.81 & \textbf{5.55 }& \textbf{6.18} & 59.73 & 36.47\\
    \bottomrule
    \end{tabular}
    \vskip -0.1in
\end{table*}

%% file: tables/instruct_table.tex
\begin{table}[t!]
\centering
\small
\caption{Experimental results on Mistral-7B-Instruct.}
\label{tab:instruct_results}
\vskip 0.15in
    \resizebox{\linewidth}{!}{
    \begin{tabular}{lccccc}
    \toprule 
    % Arena-Hard
    \multirow{2}{*}{\textbf{Model}}  &  \multicolumn{3}{c}{\textbf{Alpaca-Eval 2.0}} & \textbf{MMLU} & \textbf{GSM8K} \\
    & \textbf{LC} & \textbf{WR} & \textbf{Length} & \textbf{Acc.} & \textbf{EM} \\
    \midrule
    \multicolumn{5}{l}{\textbf{\textit{Self-Rewarding (LLM-as-Judge})}} \\
    $M_0$ & 23.52 & 16.14 & 1549 & 59.69 & 49.43 \\
    $M_1$ & 26.74 & 26.09 & 1916 & \textbf{59.79} & 50.04 \\
    $M_2$ & 24.82 & 27.20 & 2446 & 59.54 & \textbf{50.64} \\
    $M_3$ & 24.01 & 28.44 & 2871 & 59.48 & 49.81 \\
    \midrule
    \multicolumn{5}{l}{\textbf{\textit{Self-Consistent Internal Rewards (Ours)}}} \\
    $M_0$ & 22.06 & 17.64 & 1576 & 59.73 & 49.20 \\
    $M_1$ & 28.81 & 29.81 & 2122 & 59.49 & 49.51 \\
    $M_2$ & 34.83 & 34.16 & 2004 & 59.67 & 49.02 \\
    $M_3$ & \textbf{35.02} & \textbf{35.90} & 2161 & 59.59 & 48.92 \\
    \bottomrule
    \end{tabular}
    }
    \vskip -0.1in
\end{table}

%% file: tables/ablation_table.tex
\begin{table}[t]
\caption{Performance of our method on AlpacaEval after ablation.}
\label{tab:ablation}
\small
% \vskip 0.15in
 \begin{center}
 \resizebox{\linewidth}{!}{
    \begin{tabular}{lccc}
        \toprule 
        \textbf{Model} & \textbf{LC} & \textbf{WR} & \textbf{Length}\\ 
        \midrule
        Full SCIR & \textbf{24.92} & \textbf{23.85} & 1941 \\
        \midrule
        w/o Consistency Training & 22.69 & 21.61 & 1949 \\
        w/o Dynamic Preference Optimization & 22.43 & 21.49 & 1936\\
        w/o Multiple Judge Prompts& 23.84 & 23.47 & 2039 \\
        w/o Length Regularization & 23.44 & 23.72 & 2083 \\
        w/o Adaptive Reference Model & 23.58 & 22.60 & 1939 \\
        \bottomrule
    \end{tabular}
    }
\end{center}
\vskip -0.2in
\end{table}

%% file: tables/SCIR_prompt.tex
\begin{table*}[ht]
\centering
\caption{Pairwise LLM-as-a-Judge Prompts for SCIR.}
\label{tab:judge_prompts_scir}
\scalebox{0.9}{
\begin{tabular}{p{15cm}}
\toprule
\textbf{LLM-as-a-Judge Prompt 1 for Self-Consistent Internal Rewards Training}\\
\midrule
You are a highly efficient assistant, who evaluates and selects the best large language model (LLMs) based on the quality of their responses to a given instruction. This process will be used to create a leaderboard reflecting the most accurate and human-preferred answers. 
I require a leaderboard for various large language models. I'll provide you with prompts given to these models and their corresponding outputs. Your task is to assess these responses, and select the model that produces the best output from a human perspective. Here is the prompt: \\
\{ ``instruction'': ``\texttt{\{instruction\}}'', \} \\
Here are the outputs of the models: \\
\{ ``model'': ``m'', ``answer'': ``\texttt{\{response\_1\}}'' \} \\
\{ ``model'': ``M'', ``answer'': ``\texttt{\{response\_2\}}'' \} \\
Evaluate the models based on the quality and relevance of their outputs, and select the model that generated the best output. Answer by providing the model identifier of the best model. We will use your output as the name of the best model, so make sure your output only contains one of the following model identifiers: [[m]] or [[M]]. \\
\midrule
\midrule

\textbf{LLM-as-a-Judge Prompt 2 for Self-Consisent Internal Rewards Training} \\
\midrule
Please act as an impartial judge and evaluate the quality of the responses provided by two AI assistants to the user question displayed below. 
You should choose the assistant that follows the user's instructions and answers the user's question better. Your evaluation should consider factors such as the helpfulness, relevance, accuracy, depth, creativity, and level of detail of their responses. Avoid any position biases and ensure that the order in which the responses were presented does not influence your decision. Do not allow the length of the responses to influence your evaluation. Do not favor certain names of the assistants. Be as objective as possible. \\
\#\#\# Instruction: \\
\texttt{\{instruction\}} \\
\#\#\# Response m: \\
\texttt{\{response\_1\}} \\
\#\#\# Response M: \\
\texttt{\{response\_2\}} \\
Output your final verdict by strictly following this format: "[[m]]" if Response m is better than Response M, "[[M]]" if Response M is better than Response m.\\
\bottomrule
\end{tabular}
}
\end{table*}

%% file: tables/SRLM_prompt.tex
\begin{table*}[ht]
\centering
\caption{Pointwise LLM-as-a-Judge Prompt for SRLM.}
\label{tab:judge_prompts_srlm}
\scalebox{0.9}{
\begin{tabular}{p{15cm}}
\toprule
\textbf{LLM-as-a-Judge Prompt for Self-Rewarding Language Model (LLM-as-a-Judge)}\\
\midrule

Review the user's question and the corresponding response using the additive 5-point scoring system described below. Points are accumulated based on the satisfaction of each criterion: \\
- Add 1 point if the response is relevant and provides some information related to the user's inquiry, even if it is incomplete or contains some irrelevant content.\\
- Add another point if the response addresses a substantial portion of the user's question, but does not completely resolve the query or provide a direct answer.\\
- Award a third point if the response answers the basic elements of the user's question in a useful way, regardless of whether it seems to have been written by an AI Assistant or if it has elements typically found in blogs or search results.\\
- Grant a fourth point if the response is clearly written from an AI Assistant's perspective, addressing the user's question directly and comprehensively, and is well-organized and helpful, even if there is slight room for improvement in clarity, conciseness or focus.\\
- Bestow a fifth point for a response that is impeccably tailored to the user's question by an AI Assistant, without extraneous information, reflecting expert knowledge, and demonstrating a high-quality, engaging, and insightful answer.\\

User: \texttt{\{instruction\}} \\
\textless response\textgreater \texttt{\{response\}}\textless /response\textgreater \\
\\
After examining the user's instruction and the response: \\
- Briefly justify your total score, up to 100 words. \\
- Conclude with the score using the format: ``Score: \textless total points\textgreater''\\
\\
Remember to assess from the AI Assistant perspective. To evaluate the response in alignment with this additive scoring model, we'll systematically attribute points based on the outlined criteria. \\
\bottomrule
\end{tabular}
}
\end{table*}